\pdfoutput=1

\documentclass[11pt]{article}

\usepackage[final]{acl}

\usepackage{times}
\usepackage{latexsym}

\usepackage[T1]{fontenc}

\usepackage[utf8]{inputenc}

\usepackage{microtype}

\usepackage{inconsolata}

\usepackage{amsmath}
\usepackage{multirow} 
\usepackage{colortbl} 
\usepackage{graphicx} 
\usepackage{xcolor}   
\definecolor{green}{RGB}{85, 170, 85}
\usepackage{array}

\usepackage{algorithm}
\usepackage{algpseudocode}

\title{Examining and Adapting Time for Multilingual Classification via\\ Mixture of Temporal Experts}

\author{Weisi Liu  \and Guangzeng Han \and Xiaolei Huang\\
         Department of Computer Science, University of Memphis \\
        Memphis, TN, USA \\ 
         \{wliu9, ghan, xiaolei.huang\}@memphis.edu}
         
\begin{document}
\maketitle
\begin{abstract}
Time is implicitly embedded in classification process: classifiers are usually built on existing data while to be applied on future data whose distributions (e.g., label and token) may change.
However, existing state-of-the-art classification models merely consider the temporal variations and primarily focus on English corpora, which leaves  temporal studies less explored, let alone under multilingual settings.
In this study, we fill the gap by treating time as domains (e.g., 2024 vs. 2025), examining temporal effects, and developing a domain adaptation framework to generalize classifiers over time on multiple languages.
Our framework proposes \textbf{M}ixture of \textbf{T}emporal \textbf{E}xperts (\textit{MoTE}) to leverage both semantic and data distributional shifts to learn and adapt temporal trends into classification models.
Our analysis shows classification performance varies over time across different languages, and we experimentally demonstrate that MoTE can enhance classifier generalizability over temporal data shifts.
Our study provides analytic insights and addresses the need for time-aware models that perform robustly in multilingual scenarios.
\end{abstract}

\section{Introduction}



\begin{table*}[]
\begin{tabular}{c|c|c|c|c|c}
\textbf{Data} & \textbf{Language} & \textbf{Time Interval} & \textbf{Docs} & \textbf{Ave. Length} & \textbf{Imbalance-ratio} \\ \hline \hline
\multirow{4}{*}{Review} & English & 11, 12, 13, 14 & 43,835 & 68 & 33.7\\
 & French & 11, 12, 13, 14 & 24,434 & 92 & 28.0\\
 & German & 11, 12, 13, 14 & 26,803 & 117 & 43.2\\
 & Danish & 07-08, 09-10, 11-12, 13-14 & 467,215 & 111 & 26.0\\ \hline
\multirow{4}{*}{EURLEX} & English & 1958-10, 12-16& 60,000 & 1,200& 50.1\\
 & Spanish & 1958-10, 12-16& 57,785 & 1,380& 49.2\\
 & Croatian & 1958-10, 12-16& 10,444 & 1,490& 26.1\\
 & Maltese & 1958-10, 12-16& 37,521 & 1,250& 32.5\\
\end{tabular}
\caption{Overview of four time-varying languages in review corpora and legal datasets. Years from the 21st century are represented by their last two digits (e.g., 11 for 2011).  The label imbalance ratio calculates the ratio of the number of samples in the majority class/label to that in the minority class/label.}
\label{tab: dataset}
\end{table*}

Data, and therefore classification models built on the data, can change over time.
Written styles in social media platforms can change rapidly~\cite{Stewart2017Measuring,jin2024mm-soc}, and token and label distributions can shift over long periods of time~\cite{rottger2021temporal}.
Large language models~\cite{openai2024gpt4,zhao2023survey} (LLMs) pre-trained on large amount of data collections have assisted document classifiers capturing precise semantic representations of documents and achieving state-of-the-art performance on the classification task~\cite{sun2023text}.
Common ways to train and evaluate classification models depend on randomly splitting data into training, development, and test sets.
However, the training and evaluation setting rarely considers and adapts the temporal nature of classification models –– LLMs and classifiers are built on existing data while to be applied on future data subjecting to change.
Thus, a notable yet not solved question in the LLM era is: \textit{Does the temporal shifts effect classification model performance, and if so, how can we minimize the impacts?}


However, examining and encountering the impacts of temporal shifts on classification performance remain understudied. 
To estimate the impacts, studies developed classification models on news articles ~\cite{agarwal2022temporal}, legal documents~\cite{huang2019neural}, or medical notes~\cite{liu2024timemattersexaminetemporal} by comparing performance variations across time intervals of yearly or multiple years.
While retraining classifiers by new data and labels is straightforward~\cite{Rolnick2019Experience,hu2022lora,Lopez-Paz2017Gradient}, a substantial gap may prevent the method: labels of future data not be available at the time of model development.
In addition, the existing studies~\cite{rottger2021temporal, agarwal2022temporal, shang2022improving} primarily work on English corpora, little is known about if the principles are applicable to non-English languages.
Thus, a notable question yet unanswered is: \textit{can the principles of temporal shifts on English data and classifiers be applicable to other languages?}

In this study, we focus on a standard evaluation task, text classification, and answer the questions by 1) examining temporal shifts and their impacts on state-of-the-art classification models, 2) developing a \textit{domain adaptation} approach, and 3) evaluate performance impacts on the gender groups under \textbf{multilingual} settings.
Specifically, we introduce a \textbf{M}ixture of \textbf{T}emporal \textbf{E}xperts (\textit{MoTE}) by leveraging data shifts across time domains and developing a dynamic routing network to allow swiftly generalize classification models over time via two major modules: clustering-based shift evaluator and temporal router network.
We evaluate our approach on a review corpus with four languages (Danish, English, French, and German), and a legal corpus with 23 languages under a more challenging scenario with longer time span and limited training data.
We compare MoTE with state-of-the-art baselines of both non-time-adapted and time-adapted methods.
The multilingual setting has rarely been studied particularly for the temporality and its impacts on classification models.
We further conduct ablation studies to understand different MoTE modules and show that the dynamic routing mechanism plays the most critical role in promoting performance over time.
Our findings provide new insights and recommendations for both researchers and practitioners on the essential to consider time in designing and deploying models on time-varying corpora under the multilingual scenario.\footnote{Our code is available at ~\url{https://github.com/trust-nlp/TemporalLearning-MoTE}.}

\section{Data}

\begin{figure*}[htp]
\centering
\includegraphics[width=1\textwidth]{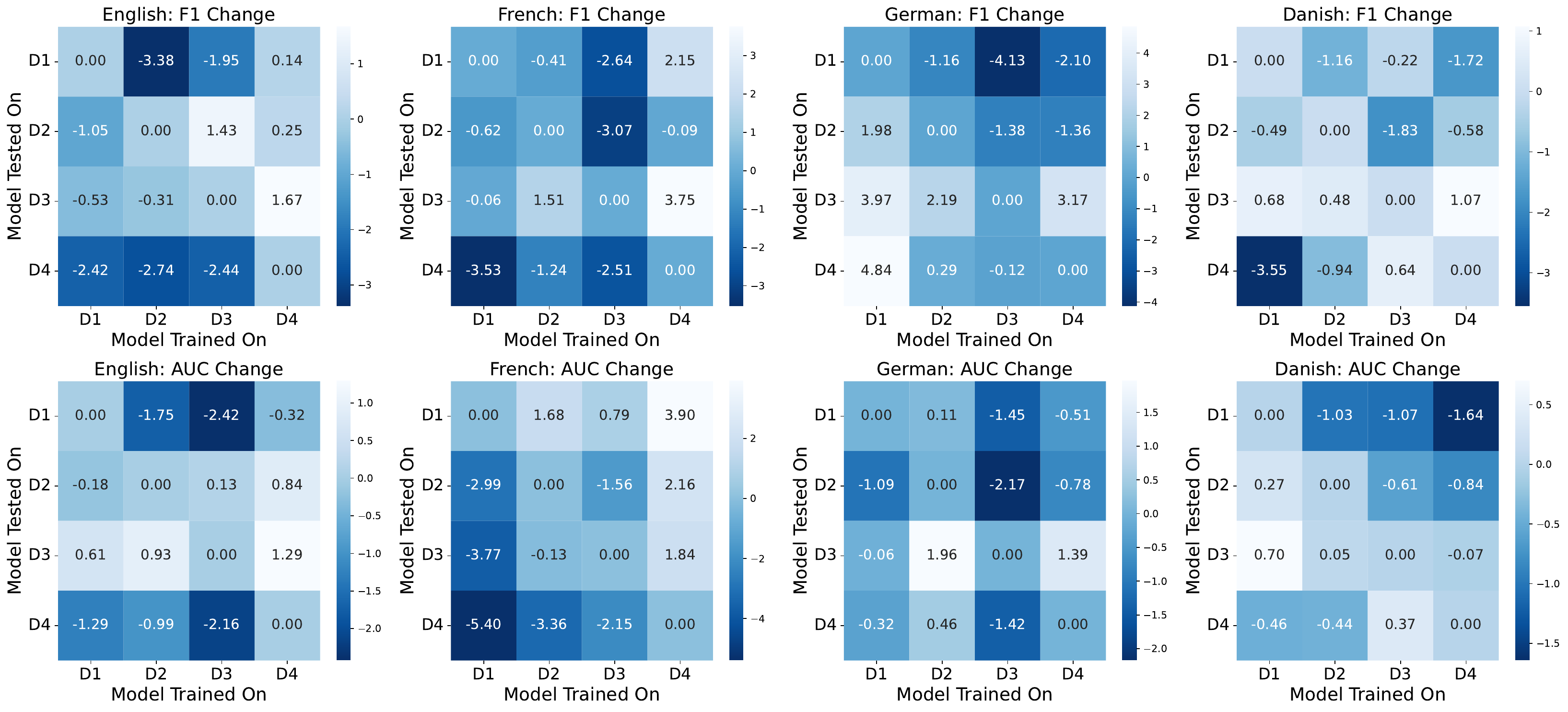} 
\caption{Visualizations of temporal effects (performance variations by macro-F1 and AUC) of the cross-domain tests on four languages. Darker blue indicate larger performance decrease.}
\label{fig: perf_change}
\vspace{-0.1in}
\end{figure*}

To better measure and understand time and its impacts on classification models, we select data from two distinct domains: user review data and legal documents. This selection enables us to compare between short, informal text (social media reviews) and longer, formal text (legal documents).
Table~\ref{tab: dataset} summarizes key statistics for the two datasets.

\paragraph{Legal Data}
Legal documents present a unique challenge due to their temporal nature, making models particularly susceptible to performance degradation over time.
The EURLEX dataset~\cite{chalkidis2021multieurlex} is a multilingual and multi-label dataset comprising legal documents from 23 European languages. 
The documents are annotated with EUROVOC concepts (labels) by the Publication Office of EU. We use the first levels of the EuroVoc
taxonomy in our experiments, encompassing 21 concepts.
Prior studies have shown that temporal effects in EURLEX significantly impact classification performance~\cite{santosh2024chronoslex}.  
We follow the original data split, with a consistent temporal division across all languages: training on documents from 1958 to 2010 and testing on documents from 2012 to 2016. 
We take English (en), Spanish (es), Croatian (hr) and Maltese (mt) as representative examples, with their statistics shown in Table~\ref{tab: dataset}. Detailed statistics for all 23 languages are provided in Table~\ref{tab: eurlex} in the appendix. 

EURLEX comprises long-form legal documents, with the average document length exceeding 1,000 words across most languages. The label imbalance ratio is calculated as the number of minority labels divided by the number of majority labels, ranges between 26.1 and 50.1 across different languages. Notably, low-resource languages, such as Croatian and Maltese, tend to exhibit lower imbalance ratios compared to higher-resource languages like English. 

\paragraph{Review Data}
we retrieve a review data ~\cite{hovy2015user} spanning years between 2007 and 2014 and covering four language corpora, Danish (da), English (en), French (fr), and German (de).\footnote{The Danish corpus ranges between 2007 and 2014, the other three range between 2011 and 2014.}
The data sources user reviews from Trustpilot website and contains 5-scale ratings, timestamps, and binary gender values (female and male).
For each corpus, we divide the data into four time intervals, remove reviewers fewer than 10 tokens, and only keep data entries with complete user gender information, which aim to ensure data quality and integrity.

The average length shows that the English reviews are generally shorter than the other corpora.
We can observe that the Danish data is much larger than the other languages, due to the origin of the Trustpilot website.
The The label imbalance ratio calculates the ratio of the number of samples in the majority class to that in the minority class. 
Since the review data also contains user gender information, we additionally report the gender imbalance ratio, which is calculated as the proportion of the majority gender to the minority gender. The gender imbalance ratios for the review data are 1.51 (en), 1.29 (fr), 1.29 (de), and 1.14 (da)
We can observe that both label and gender distributions are not balanced: majority ratings can be as twice as minority labels, and there are more male than female users.
To properly estimate temporal effects, we consider the imbalance factor in our evaluation metrics for classification performance.

\subsection*{Multilingual Analysis of Temporal Effects}
We take the review data as an example to explore the temporal effect. 
We define the \textbf{temporal effect} as existing when there is a \textit{performance variation} in at least one time domain pairs, measured by the difference between cross-domain and in-domain performances.
For the cross-domain performance $p_{ij}$, we train a model on the training set of the time domain $D^i$ (source) and evaluate the test set of the time domain $D^j$ (target). For the in-domain performance $p_{ii}$, we train and test a model on the training and test sets from the same time domain $D^i$.

To assess temporal effect under multilingual settings, we split each corpus into $|T|$ ($T=4$ in this study) splits and treat each time interval as a domain. 
Therefore, we can represent each data per time domain as $D^t (t \in T)$. 
We sampled down the data size for each time domain to match the domain with the smallest size to mitigate systematic errors due to the data size. 
For each time domain $D^t$, we split the data by 7:3 for training and testing to study temporal effect. 
We use macro averaged F1 score (F1-ma) and AUC to measure in-time-domain and cross-time-domain performances of classifiers.
To measure temporal effects on the domain pair ${D^i, D^j}$, we subtract the in-domain performance $p_{jj}$ ($D^j \rightarrow D^j$) from the cross-domain performance $p_{ij}$ ($D^i \rightarrow D^j$), i.e. the \textit{performance change} from the source domain $D^i$ to the target domain $D^j$. 
Each experiment repeats three times, takes average of the performance, and estimates temporal effects by the performance variations ($p_{ij}-p_{jj}$).
Finally, we visualize temporal effects as performance variations across time domains in Figure~\ref{fig: perf_change}.

Figure~\ref{fig: perf_change} suggests that classifiers perform worse in the other time domains and generally diminishes more performance with a longer time interval, while different languages exhibit different but similar temporal patterns.
For example, English, French, and Danish classifiers drop more than 3\% when training on D1 and testing on D4; and close time domains of all four languages usually have smaller performance variations.
The consistent observations can suggest that: 1) temporal effects exist and diminish on classification performance in the multilingual scenario; and 2) temporal patterns can vary across different languages.
The findings inspire us to propose the Mixture of Temporal Experts (MoTE) to learn and adapt the time into classifiers.

\section{Mixture of Temporal Experts (MoTE)}

\begin{figure*}[htp]
\centering
\includegraphics[width=0.9\textwidth]{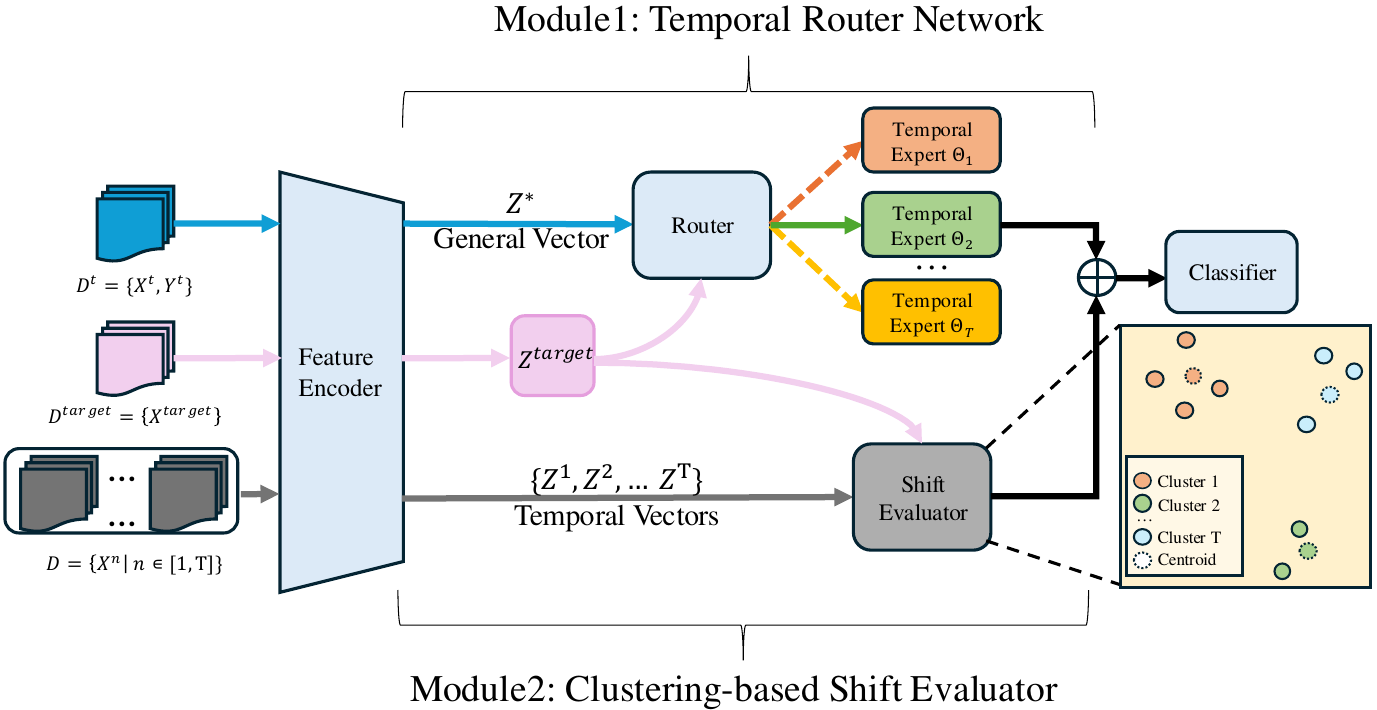} 
\caption{The MoTE method overview. $D^t$ is in the source time domain data that has true labels, $D$ is temporal ordered data excluding target domain, and $D^{target}$ is the target time domain data without labels. Blue and grey lines indicate the training process, and pink line represent predicting data flow in the target time domain.}
\label{fig:mote}
\vspace{-0.1in}
\end{figure*}

In this section, we introduce a multi-source domain adaptation approach, Mixture of Temporal Experts (MoTE) in Figure~\ref{fig:mote}, which leverages temporal data shift and mixture of temporal experts. 
The MoTE method contains two modules: 
Clustering-based Shift Evaluator and Temporal Router Network.
The Clustering-based Shift Evaluator measures data shift by the difference between the data representation of input data and the established cluster centroids. 
The Temporal Router Network module consists of a router and multiple experts. The router is trained to dispatch the data representations to experts based on it's distance to the cluster centroids. 
And then the experts integrate the data shift vectors produced by the shift evaluator before passing to the classification layer.
Finally, the predictor will take the time-aware representations for predictions.

\paragraph{Problem Definition}
For each language corpus, we hold out the most recent data as target time domain data $D^{target}$ and the rest as source time domain $D^{source}$, and divide $D^{source}$ to $T$ time domains by time order: $D^t, t \in \{1,...,T\}$.
The source model $\Theta_{source}$ is the model fine-tuned on source domain data.
The performance decline when the source model $\Theta_{source}$ is tested on the target domain data $D^{target}$ indicates that the source model struggles to generalize effectively to the target data, and thus requires adaptation method to mitigate performance degradation.

Our goal is to develop a method that can adapt $\Theta_{source}$ to the target domain, which can be aware of the data shifts between $D^{source}$ and $D^{target}$ and improve model's performance on the target data, in which we achieve by the two major modules, clustering-based shift evaluator and temporal router network. 

\subsection{Module 1: Clustering-based Shift Evaluator}
The Clustering-based Shift Evaluator module quantifies data shift in text data over time to address model performance degradation. 
It creates semantic representations of text using a feature encoder, then applies K-Means clustering to these representations. 
The module labels source data with the nearest clusters, creating a warmup dataset for expert routing.

To represent data shift, the module calculates the distance between target data and source cluster centroids. 
This distance serves as a shift representation, informing experts about how new data differs from the original distribution. 
By quantifying shift in this semantic space, the module enables the system to adapt to evolving data distributions.
A detailed pseudocode is provided in Appendix~\ref{sec:appendix:psudocode}, shown in Algorithm~\ref{alg:shift_evaluator}.

\paragraph{Data Clustering}

Data shift is a key factor contributing to model performance degradation. Unlike structured data, which can be directly quantified for data shifts using feature-based methods, text data lacks such explicit features.
Therefore, we encode the text and use embeddings to quantify the data shift in a semantic space.
We use the output of feature encoder as the data semantic representation. 
Given time domain data $D=\{X^{1},X^{2}, ..., X^{t}\}$ the data representation encoded by feature encoder are denoted as: $Z=\{Z^{1}, Z^{2}, ..., Z^{t}\}$, where D is temporal ordered data exclude target domain.
We run a K-Means clustering algorithm on the representations, with the number of clusters set to $T$, matching the number of time domains. 
Our experiments also tried other clustering algorithms (e.g., Spectral Clustering), while K-Means achieved the fastest parameter convergence and the best performance.
The $T$ source-time data clusters are denoted as $C_1, C_2,..., C_T= \text{\textit{Kmeans}}(Z, T)$, with $T$ cluster centroids: $\{c_{1}, c_{2}, ..., c_{T}\}$.

We label the source data $z_i$ with the nearest cluster $l_i$ 
and save the clustered data $(Z, L)=\{(z_i,l_i)\}$ as a warmup dataset for the mixture of experts' routing process. 
The primary purpose of retaining these cluster labels is to automatically segregate data with different distributions and match it to the nearest cluster. 
This warmup process allows the router to learn how to dispatch incoming data to the appropriate expert by identifying the most relevant cluster.
Additionally, the cluster centroids are preserved to calculate data shift representations.

\paragraph{Data Shift Representation} 

The data clusters is based on the semantic distribution of the data.
The distance between the target data and the cluster centroids of source data can be a measurement for the data shift between the source and target data.
We use the difference between target data and centroids of the source data clusters as a \textit{data shift representation}, and this representation is provided to the experts as a data shift information.

For $z^{target}_{i} \in Z^{target},$ $v_{ij}=z^{target}_{i}-c_{j}, j\in {1,2,...,T}.$ $v_{ij}$ is later passed to expert $\Theta_j$ if the router dispatch representation $z^{target}_{i}$ to the expert $\Theta_j$ , indicating it's shift from the $j^{th}$  cluster centroid. 

\subsection{Module 2: Temporal Router Network}

The Temporal Router Network adapts to data shifts by routing inputs to appropriate experts. 
It consist of a mixture of $T$ expert networks: $\Theta_1 , \Theta_2, ..., \Theta_T$ and a router. 
Each expert is a transformer block~\cite{Attention2017Vaswani} with a classification layer, and the router is a gating network $G$ whose output is a $T$-dimension routing vector. 
The module uses a top-K gating mechanism, integrates data shift information, warms up using clustered source data, and produces final predictions by weighted averaging of expert outputs.
A detailed pseudocode is provided in Appendix~\ref{sec:appendix:psudocode}, shown in Algorithm~\ref{alg:router_network}.

\paragraph{Routing}
The Router is a top-K gating network, with a trainable linear network with weight $W_g$ and with softmax and  top-K gating function. 
Given a input token $z$, the routing vector is:
\begin{equation*}
{G}(z)=\text { \textit{TopK }}(softmax(W_g\cdot z),K)
\end{equation*}
\begin{equation*}
\text {\textit{TopK }}(v, K)_i= \begin{cases}v_i & \text{if } v_i \in \text { top } K \text {elements}  \\ 0 & \text { otherwise. }\end{cases}
\end{equation*} 

To prevent the gating network fall into state that always produces large weights for the same few experts, we added a auxiliary load-balancing loss $L_{aux}$~\cite{shazeer2017outrageously} to Cross-entropy loss:


\begin{equation*}
L = \text{CE}(y, \hat{y}) + \lambda L_{\text{aux}},
\end{equation*}

where $\text{CE}(y, \hat{y})$ represents the cross-entropy loss, $y$ is the true label, and $\hat{y}$ is the predicted probability distribution over the classes. For all experiments, we set $\lambda = 0.01$, following the original setting~\cite{shazeer2017outrageously}.

\paragraph{Warmup}

We use the clustered source data representations $(Z, L)$ to warm up the router. Each representation is labeled with a cluster index obtained during the data clustering process. 
Intuitively, this process is to ensure that each expert corresponds to a specific cluster, allowing them to learn how to handle data from their respective clusters more effectively during subsequent training. 
In the warmup stage, we train the router to dispatch data that is closer to a particular cluster in the latent semantic space to the corresponding expert.

\begin{table*}[htp]
\centering
\resizebox{1\textwidth}{!}{
\begin{tabular}{c||cccccccccccc}

             \multirow{2}{*}{Method} & \multicolumn{3}{c}{English} & \multicolumn{3}{c}{French} & \multicolumn{3}{c}{German} & \multicolumn{3}{c}{Danish} \\
             
              & F1-ma    & AUC-ma      & Fair$\downarrow$  & F1-ma    & AUC-ma      & Fair$\downarrow$ & F1-ma    & AUC-ma      & Fair$\downarrow$ & F1-ma    & AUC-ma      & Fair$\downarrow$ \\

\hline
ChronosLex & 47.96 & 79.31 &  2.23 &  41.48 &  76.36 &  2.04 &  38.53 &  75.12 &  2.17 &  47.64 &  82.01 &  2.16 \\
  
Anti-CF       & 25.31  & 65.09  & 2.14  & 15.63  & 65.59  & 1.80 & 16.78  & 55.61  & 1.99 & 32.77  & 71.58  & 2.11 \\

Self Labeling & 40.52  & 72.02  & 2.13  & 38.07  & 75.57  & 1.97 & 28.30  & 69.42  & 2.08 & 46.51  & 77.95  & 2.03 \\ 

Source Model  & 48.92  & 78.64  & 2.21  & 35.05  & 72.25  & 2.00 & 34.01  & 75.06  & 2.15 & 49.83  & \textbf{82.58}  & 2.13 \\

MoTE (ours) &  \textbf{53.30} &  \textbf{80.81} & 2.17 &  \textbf{44.97} &  \textbf{77.24} &  1.97 &  \textbf{46.70} &  \textbf{76.29} &  2.14 &  \textbf{52.77} &  82.57 & 2.06 \\

\hline
$\Delta$ (MoTE - Avg) & +12.62 & +7.05 & -0.01 & +12.41 & +4.80 & +0.02 & +17.30 & +7.49 & +0.04 & +8.58 & +4.04 & -0.05 \\

\end{tabular}
}
\caption{Overall classification performance. The row $\Delta$ represents the difference between MoTE and the average of all baselines for each metric. Positive values indicate improvement, while negative values indicate performance decreases. For Fair, negative values indicate worse performance as lower is better.}
\label{experiments}
\vspace{-0.1in}
\end{table*}

\paragraph{Data Shift Information Integration} 
The feature encoder produces general vectors
 $Z^*=\{z_1^*,z_2^*,....,z_n^*\}$
 where $z_i^*=\text{\textit{Encoder}}(x_i)$.
 The router produce a gating vector $G(z_i^*)$
and dispatch $z_i^*$ to K experts. 
shift vector $v_{ij}=z_i-c_j$
We concatenate the shift vector $v_{ij}$ and the $j^{th}$ expert's transformer layer's [cls] token output:
$E(z_i^*,\Theta_j)\oplus v_{ij}$ before pass it to classification layer.

\paragraph{Weighted Averaging}
The MoTE model produce a set of probabilities $p_k$ from each experts for the target data prediction. 
The final output of the adapted model is obtained by passing these predictions through a weighted averaging function and a sigmoid function, where the weights are the routing scores from the top-K gated router.
\begin{equation*}
    p(x,K)= \frac{1}{|K|} \sum_{k\in T} G_{k}(z^*)*p_{k}(x)
    \label{averaged_logits}
\end{equation*}

Where T is the number of experts and $G(z^*)$ is the gating vector from the router.

\begin{table*}[]
\centering
\resizebox{1\textwidth}{!}{
\begin{tabular}{ll|ccc|ccc|ccc}
Data Summary &     & \multicolumn{3}{c|}{SOTA Baseline (\%)} & \multicolumn{3}{c|}{MoTE Performance (\%)} & \multicolumn{3}{c}{ $\Delta$(MoTE-SOTA) (\%)} \\ 
Language     & ISO & AUC-ma       & F1-sa      & F1-ma      & AUC-ma        & F1-sa       & F1-ma       & AUC-ma           & F1-sa          & F1-ma          \\ \hline \hline
English    & en & 87.53 & 68.77 & 43.49 & 88.04 & 72.95 & 57.52 & 0.51 & 4.18  & 14.03 \\
German     & de & 86.78 & 68.12 & 40.99 & 88.63 & 71.55 & 54.54 & 1.85 & 3.43  & 13.55 \\
French     & fr & 87.19 & 64.84 & 37.09 & 88.14 & 71.82 & 54.73 & 0.95 & 6.98  & 17.64 \\
Italian    & it & 87.59 & 69.53 & 44.2  & 88.15 & 72.65 & 56.09 & 0.56 & 3.12  & 11.89 \\
Spanish    & es & 87.28 & 68.31 & 42.93 & 88.63 & 72.73 & 57.65 & 1.35 & 4.42  & 14.72 \\
Polish     & pl & 87.18 & 68.35 & 40.35 & 88.27 & 72.00    & 54.75 & 1.09 & 3.65  & 14.40  \\
Romanian   & ro & 88.07 & 70.63 & 45.6  & 89.27 & 73.51 & 56.62 & 1.20  & 2.88  & 11.02 \\
Dutch      & nl & 87.11 & 69.69 & 43.99 & 87.76 & 71.76 & 55.03 & 0.65 & 2.07  & 11.04 \\
Greek      & el & 86.37 & 66.89 & 38.98 & 87.13 & 69.17 & 51.92 & 0.76 & 2.28  & 12.94 \\
Hungarian  & hu & 87.28 & 66.89 & 39.65 & 87.93 & 71.05 & 53.62 & 0.65 & 4.16  & 13.97 \\
Portuguese & pt & 89.99 & 74.52 & 52.61 & 89.39 & 74.70  & 59.4  & -0.60 & 0.18  & 6.79  \\
Czech      & cs & 87.29 & 68.45 & 41.39 & 88.20  & 72.65 & 54.18 & 0.91 & 4.20   & 12.79 \\
Swedish    & sv & 86.55 & 65.08 & 37.97 & 88.68 & 71.52 & 54.03 & 2.13 & 6.44  & 16.06 \\
Bulgarian  & bg & 88.78 & 71.48 & 48.51 & 89.53 & 74.60  & 61.07 & 0.75 & 3.12  & 12.56 \\
Danish     & da & 87.50  & 69.52 & 42.93 & 88.36 & 72.47 & 56.04 & 0.86 & 2.95  & 13.11 \\
Finnish    & fi & 85.02 & 59.46 & 31.54 & 86.57 & 70.64 & 52.3  & 1.55 & 11.18 & 20.76 \\
Slovak     & sk & 88.57 & 71.68 & 48.48 & 89.3  & 71.99 & 56.55 & 0.73 & 0.31  & 8.07  \\
Lithuanian & lt & 87.15 & 68.13 & 41.87 & 88.19 & 72.41 & 56.5  & 1.04 & 4.28  & 14.63 \\
Croatian   & hr & 87.64 & 69.55 & 46.45 & 88.56 & 69.74 & 54.62 & 0.92 & 0.19  & 8.17  \\
Slovene    & sl & 87.63 & 69.43 & 42.98 & 89.14 & 72.30  & 56.27 & 1.51 & 2.87  & 13.29 \\
Estonian   & et & 87.12 & 68.16 & 40.28 & 87.40  & 71.73 & 54.63 & 0.28 & 3.57  & 14.35 \\
Latvian    & lv & 87.58 & 68.88 & 41.94 & 88.71 & 70.87 & 55.17 & 1.13 & 1.99  & 13.23 \\
Maltese    & mt & 82.67 & 56.36 & 32.25 & 84.61 & 61.51 & 44.15 & 1.94 & 5.15  & 11.9 
\end{tabular}
}
\caption{Comparison of MoTE performance and state-of-the-art baseline on 23 language splits EURLEX data. $\Delta$ measures the MoTE's absoluate percentages of performance improvements over the SOTA baseline.}
\label{tab: eurlex_result}
\end{table*}
\section{Experiments}

To examine the effectiveness of our proposed Mixture of Temporal Experts (MoTE) method, we conduct experiments to compare our method with the source model (fine-tuned on source time domain) and state-of-the-art baselines on the target time domain data.  
We set our experimental result against three established baseline methods: Self-Labeling~\cite{agarwal2022temporal}, Anti-CF~\cite{su2023beware}, and ChronosLex~\cite{santosh2024chronoslex}. 
We set the source model as a comparison to see whether the method improves the model's adaptation to the target domain.
Our assessment includes both performance evaluation and fairness evaluation.
To validate domain-generalizability of our MoTE method under multi-lingual setting, we also conduct experiments on the legal data EURLEX~\cite{chalkidis2021multieurlex} containing 23 languages and compare our method with the best-performing baseline~\cite{santosh2024chronoslex}.
In addition, our work includes an ablation study focusing on the proposed MoTE method, where we removed each component of our method.
By assessing the performance resulting from the removal of each component, we measured its contribution to the overall effectiveness.


\subsection{Baselines}
To demonstrate the effectiveness of our proposed MoTE method, we compare it with the source model and three state-of-the-art baselines.
\textit{Source Model} is the vanilla baseline for temporal domain adaptation.
We obtain the model fine-tuned on gold labeled source data, and then test this source model on target data.
\textit{Self-Labeling}~\cite{agarwal2022temporal} employ the fine-tuned source model to generate self-labeled target data, which is then combined with the original source data to train a new model.
\textit{Anti-CF}~\cite{su2023beware} utilizes the source model's output to regularize model updates and includes a sequence of adapters attached to the frozen source model for efficient inference.
\textit{ChronosLex}~\cite{santosh2024chronoslex} train the model on data from each time period sequentially, moving chronologically through the temporal splits.

\subsection{Evaluation Metrics}
Our evaluation employed F1 scores and area under the ROC curve (AUC). 
The review data have user demographic information, so we also include a fairness metric to assess the fairness over different demographic groups for a more holistic comparison of our approach against the baselines. 
The fairness metric is measured by the \textit{equality differences} of false positive/negative rates (FPR/FNR)~\cite{dixon2018measuring}, which calculates the absolute discrepancies in FPR and FNR across different gender groups. 
For example, the equality difference of FPR is calculated by $\sum_{g\in G}|FPR_g-FPR|$, where $G$ is the gender and $g$ is a gender group (e.g., female).
We report the sum of the equality differences of FPR and FNR, denoted as “Fair”.
For the experiment on EURLEX data, since the task is multi-label classification and the data lack of subgroup information for fairness evaluation, we report samples-F1, macro-F1 score and AUC.

\subsection{Experimental Settings}
For the review data, we hold out the 20\% data of the most recent time period from each corpus as the target time domain, with the remaining data used as the source time domain.
For the legal data, we follows the train-test split of the EURLEX dataset and use the training split as the source time domain.
We use \textit{XLM-RoBERTa-base}~\cite{xlm-roberta} as the base model and feature encoder in our experiments, with learning rate 3e-5, batch size 32, and a maximum token length of 128 for the review data and 512 for the legal documents. with the early stopping to tune for best performance.
All hyperparameters and baseline experiments follow their original settings, with details provided in Appendix~\ref{sec:appendix:implementation}.


\section{Results}
We present the macro-F1 score and AUC along with fairness evaluation of the review data in Table \ref{experiments}. 
And the results of legal data are presented in Table \ref{tab: eurlex_result}. 
To better interpret the experimental results, we conduct a detailed performance analysis.

\paragraph{Analysis 1. Does MoTE improves model's generalizability?}

We evaluate our proposed MoTE method across two distinct domains: user reviews (short, informal texts) and legal documents (long, formal texts). The results consistently demonstrate MoTE’s ability to enhance model performance over time.
The result on the review data in Table \ref{experiments} shows that our proposed MoTE method has better F1 score by a range of 
2.94\% to 12.69\% improvement compare to the source model.
This shows that our proposed MoTE method significantly improves the  model's generalizability over time.
Besides, our proposed \textit{MoTE method consistently performs better than baselines} across all four languages and all metrics: MoTE has better F1 score by a range of 5.13\% to 8.17\% improvement compare to the best performing baseline (i.e. ChronosLex).
On the legal data, as shown in in Table \ref{tab: eurlex_result}, MoTE achieves improvements in macro-averaged F1 score ranging from 6.79\% to 17.64\% and in samples-averaged F1 score ranging from 0.18\% to 11.18\% compared to the SOTA baseline (ChronosLex)
These improvements shows MoTE’s effectiveness in handling longer texts and more structured, formal language, which often span longer time intervals.

\paragraph{Analysis 2. Does MoTE improves fairness?}
As shown in Table \ref{experiments}, our proposed MoTE method has better group fairness compared to the source model across all languages' result(e.g. improves by 3.28\% in Danish corpus). 
Although certain baselines, such as Anti-CF and Self-Labeling, achieve slightly better fairness scores, they suffer from significantly lower overall performance in both F1 score and AUC.
Overall, our method enhances fairness without sacrificing performance improvement.


\paragraph{Analysis 3. What's the improvement difference across different languages?}

We find that our method has best performance improvement in German corpus within the review data (F1 improved 12.69\% over the source model) and best group fairness improvement in Danish corpus (Fair improved 3.28\%).
The German corpus is the most class-imbalanced corpus (largest imbalance-ratio in Table \ref{tab: dataset}), with the number of samples in majority class is 43.2 times greater than that in minority class.
And the Danish corpus is the most gender-balanced corpus, with lowest gender-imbalance ratio of 1.14. 
In the legal domain, MoTE also shows substantial improvements in low-resource languages, such as Maltese, where F1-ma improves by 11.9 percentage points, composes a relative increase of 36.9\%.
These may indicate:

1) MoTE is particularly effective for class-imbalanced datasets (e.g. German corpus).
The use of a mixture of temporal experts enhances predictions for minority classes, which are more impactful in the macro-F1 score.

2) MoTE improves fairness more in gender-balanced datasets (e.g. Danish corpus). This may be because a more balanced demographic distribution offers more minority group data, enabling MoTE to better address fairness.

\paragraph{Analysis 4. Why Anti-CF and Self-Labeling failed?}

We observe that methods like Anti-CF and Self-Labeling lead to large degrees of performance decline compare to source model, which means the adaptation failed. 
These approaches depend heavily on the source model's outputs on the target data: 
Anti-CF uses the source model’s output to regularize the adapted model, while Self-Labeling relies on the source model's labels on the target data to train the new model. 
A possible explanation for this performance decline is that such reliance can introduce noise, particularly when there is substantial data shift in the minority classes. 
In this case, the source model's incorrect predictions may amplify errors in these classes, leading to reduced macro-F1 and macro-AUC scores. This could explain why the performance decline is more pronounced in language corpora where the source model already exhibits lower macro-F1 and macro-AUC scores, such as French and German.



\begin{table*}[htp]
\centering
\resizebox{1\textwidth}{!}{
\begin{tabular}{c||cccccccccccc}
\multirow{2}{*}{Method}& \multicolumn{3}{c}{English} & \multicolumn{3}{c}{French} & \multicolumn{3}{c}{German} & \multicolumn{3}{c}{Danish}\\
& F1-ma  & AUC-ma    & Fair$\downarrow$ & F1-ma  & AUC-ma    & Fair$\downarrow$ & F1-ma & AUC-ma & Fair$\downarrow$ & F1-ma  & AUC-ma & Fair$\downarrow$ \\ \hline
w/o warmup & 52.85 & 76.75 & 2.19 & 37.73 & 72.75 & 1.89 & 45.40 & 72.60 & 2.12 & 49.15 & 80.13 & 2.09 \\
w/o router & 40.66 & 73.64 & 2.18 & 40.20 & 71.60 & 1.96 & 42.53 & 70.95 & 2.16 & 49.93 & 79.43 & 2.10 \\
w/o evaluator & 47.56 & 79.87 & 2.18 & 34.83 & 68.92 & 1.97 & 44.77 & 72.73 & 2.14 & 48.78 & 77.82 & 2.06 \\
MoTE &
  \textbf{53.30} &
  \textbf{80.81} &
   2.17 &
  \textbf{44.97} &
  \textbf{77.24} &
  1.97 &
  \textbf{46.70} &
  \textbf{76.29} &
  2.14 &
  \textbf{52.77} &
  82.57 &
  2.06 \\
\end{tabular}
}
\caption{Ablation analysis results. We bolden the highest F1 and AUC scores. $\downarrow$~indicates lower is better.}
\label{ablation}
\vspace{-0.1in}
\end{table*}

\subsection{Ablation Study}


To systematically explore the contribution of each component in our proposed MoTE method, we conducted an ablation study by removing each key components from the proposed MoTE model: the warmup process, the router network, and the shift evaluator.
Removing the first two components may lead the data fail to be routed to the ideal expert, and missing the shift evaluator may result in the model unaware of data shift.
We summarized the result of the ablation study in Table \ref{ablation}.

1) \textbf{w/o warmup:} removed the warmup process for the router.  Removing the warmup process may cause overfitting issue on a special expert.

2) \textbf{w/o router:} removed the routing network, instead we randomly dispatch the data to one single expert. In this case, all the experts may underfit the data because the sparsity of data.

3) \textbf{w/o evaluator} removed the clustering-based shift evaluator, with no temporal data shift information provided to the temporal experts.

\paragraph{Difference in performance decline by removing component} We find that \textit{removing each module all results in different degrees of performance decline} (e.g. macro-F1 significantly declines by a range of 4.77\% to 10.14\% for French corpora's result). 
The components causing the most substantial performance drop differ across datasets. 
For English and German corpora, removing the router (w/o router) has the largest impact, while for French and Danish corpora, removing the evaluator (w/o evaluator) results in the most significant decline. 
The largest performance degradation occurs in the French corpora when removing the shift evaluator, with a 10.14\% decrease compared to the complete MoTE method.
This observation indicates the importance of both the mixture of experts architecture and the data shift information in the MoTE method. The warm-up process, while still beneficial, appears to have a relatively smaller impact on overall performance. 

\paragraph{Difference in ablation result of four languages} We observe that while performance declines occur across all datasets when removing any component, the magnitude of these declines is notably smaller for English and Danish. 
We infer that other factors (e.g. document length and class-imbalance) influence the performance diminish in ablations.

In conclusion, each component contributes to the temporal learning effectiveness, with the complete MoTE model achieving the best overall performance and relatively lower algorithmic biases over the gender.

\section{Conclusion}

Our study suggests that time can significantly impact classification performance and gender fairness under the multilingual setting, and model deployments should be aware of the temporal effects may vary across different languages and gender groups.
The proposed MoTE approach outperform the state-of-the-art classifiers and time-aware baselines by a large margin and maintain gender fairness, as demonstrated both in review and lagal domain. 
Our ablation analysis suggests that while the temporal expert and shifting evaluator modules both contribute to the performance improvements, the temporal expert module acts a more critical role, verifying our initial motivation.
Finally, the findings also provide a recommendation for practitioners following the insights above that it is important to slice test entries from future or new samples of any time-varying data, instead of only randomly slicing samples as a test set.

\section{Acknowledgment}
We would like to thank anonymous reviewers for their valuable comments.
This project is partly supported by the NSF (National Science Foundation) IIS-2245920. 
The first two authors are supported mainly by the NSF CNS-2318210 and partially by generous contributions from the College of Arts and Sciences and Information Technology Services at the University of Memphis.
The computing resources from the iTiger GPU cluster\footnote{\url{https://itiger-cluster.github.io/}, funded by the NSF CNS-2318210.} primarily supported the experiments.

\section{Limitations}

While we have examined temporal effects and adapted the time into the classification models by the MoTE approach under the multilingual setting, two major limitations have be acknowledged to appropriately interpret our findings.
First, we conducted our experiments on the review data and legal data, while the observations may vary across other fields, such as medical and clinical data.
Expanding our approach may require complete information of timestamp, language categories, and demographic attributes (e.g., gender). \textbf{However}, the complete data entries are usually hard to collect.
In this study, we include four languages in review data and evaluate different approaches over four time intervals to demonstrate how temporal effects can impact algorithmic classifiers, we further propose a MoTE approach and test it both on review and legal data, expanding the experiments to 23 languages.
Second, we empirically selected the XLM-RoBERTa-base~\cite{xlm-roberta} as our neural feature encoders to extract vector representations of documents, and other language models may be alternatives.
Performance improvements by the other language models may vary during real deployment and practice.
In this study, we chose the base model as the it achieved the best performance in the classification tasks~\cite{xlm-roberta} and the baselines.
To enhance fair comparisons, we kept the same base model as our baselines in this study, while alternative language models will be evaluated in our future studies.
Third, we created four temporal domains by averaging document distribution across different years for the review data. However, we did not explore alternative domain-splitting strategies. Different domain-splitting methods (e.g., unequal time intervals or document-type-based partitions) could potentially yield varying insights into temporal effects and adaptation performance. Future studies could evaluate such alternative strategies to generalize our findings.

\bibliography{mote}

\appendix


\section{Related Work}

\subsection{Model Generalizability}

Maintaining model performance across various scenarios and settings is an essential challenge in machine learning. Model generalizability is a broad concept that encompasses the ability to handle challenges such as data imbalance~\cite{jones2024examining}, varying data sizes~\cite{jin2022prototypical}, and shifts in data distributions~\cite{su2023beware}.
Time is a critical yet under-explored aspect of model generalizability. Recent research on time as a factor in machine learning models has largely focused on a model’s temporal awareness, such as temporal reasoning~\cite{xiong-etal-2024-large,Yang2024Temporal} and time prediction~\cite{xiong_2024_TEILP}, temporal generalizability—ensuring that models maintain performance under time-evolving data distributions—remains less explored. 
In this study, we fill this gap by treating time as a domain and developing a domain adaptation based approach called \textbf{M}ixture of \textbf{T}emporal \textbf{E}xperts (\textit{MoTE}), to dynamically adapt models to data temporal shifts.

\subsection{Domain Adaptation}


Domain adaptation (DA)~\cite{daume2006domain,blitzer2006domain,ben2010theory,farahani2021brief} is a set of model optimization and data augmentation methods to promote model performance, assuming that data distributions change between training and test steps.
DA has several major directions to improve text classification robustness, including pivot features~\cite{blitzer2006domain,li2022cross}, instance weighting~\cite{jiang2007instance,Lv2023Review}, and domain adversaries~\cite{Ganin2015Unsupervised,Kong2024Unsupervised,zeng2024Unsupervised}. 
However, very few studies have treated time as domains and developed new domain adaptation methods to model temporal shifts.
Our study treats time as domains and develops a multi-source adaptation approach, \textit{MoTE}, to learn time and promote model generalizability.


Previous works have considered adapting temporal effects in text classifiers, such as continuously pre-training language models~\cite{rottger2021temporal, agarwal2022temporal, shang2022improving} and diachronic word embeddings~\cite{huang2019neural, rajaby2021time,dhingra2022time}. 
However, domain adaptation has not been fully explored in those studies.
Several recent works have employed domain adaptation to address temporal shifts~\cite{he2023domain,ott2022domain} on structured data (sensor data) by the pivot feature approach that sets the feature space of the target domain as a pivot and aligns feature vectors of the source domain towards the pivot.
However, such approaches may not be applicable to the unstructured text data, which has high dimensional features and sequential dependencies –– the focus of our study.

In contrast, our study proposes a multi-source domain adaptation approach (MoTE) to model temporal effects into classification models. Particularly, the existing studies primarily focus on English data leaving multilingual classification scenarios underexplored, which has been examined in our study.

\subsection{Multilingual Classification}

%

The remarkable success of language models has led to significant advancements in multilingual text classification, addressing challenges such as multilingual long-text classification~\cite{chalkidis2023chatgpt} and parameter-efficient multilingual classification~\cite{razuvayevskaya2024comparison}.
For example, in order to solve the challenge of multilingual long text classification,~\cite{chalkidis2022lexglue} uses a hierarchical attention mechanism to improve the context window of pre-trained language models. 
Additionally, researchers have leveraged the advanced capabilities of multilingual LLMs~\cite{xue2021mt5,ma2021contributions} for multilingual classification in both few-shot~\cite{wang2020Generalizing} and zero-shot ~\cite{yin2019benchmarking} settings.
As an instance, recent studies have evaluated the performance of ChatGPT~\cite{lai2023chatgpt} and m-GPT~\cite{shliazhko2023mgpt} on multilingual text classification tasks in a zero-shot setting, demonstrating the generalization capability of these large language models on unseen multilingual datasets.

However, due to the significant disparity between English and other languages in high-quality corpus data, existing state-of-the-art approaches typically leverage the language model's English capability to enhance classification in other languages, as exemplified by machine translation-augmented text classification~\cite{king2024using} and cross-lingual in-context learning~\cite{cueva2024adaptive}. 
Consequently, recent studies only focus on the temporal adaptation of English classification task~\cite{agarwal2022temporal,dhingra2022time}, the impact of temporal shifts and trends in non-English languages on multilingual text classification performance remains largely unexplored.
In contrast, our work propose the Mixture of Temporal Experts (MoTE), aims to investigate the temporal shifts in multilingual text classification data.

\section{Algorithm Details}
\label{sec:appendix:psudocode}
We provide the pseudocode of the proposed MoTE architecture, which consists of two key modules: the Clustering-Based Shift Evaluator and the Temporal Router Network. 
Both modules operate on top of a frozen feature encoder, which extracts data representations from the input. 
The Clustering-Based Shift Evaluator module evaluates shifts in data distributions using historical data representations and computes cluster-based shift vectors, as described in Algorithm~\ref{alg:shift_evaluator}.
The Temporal Router Network then uses these shift vectors and warmup data to train a temporal router network that dynamically assigns experts to different data representations, as shown in Algorithm~\ref{alg:router_network}.

\begin{algorithm*}[ht]
\caption{Clustering-Based Shift Evaluator}
\label{alg:shift_evaluator}
\begin{algorithmic}[1]
\Require $T$ (Number of clusters), $Z^{\text{source}}$ (Historical data), $Z$ (Current data)
\Ensure $\text{WarmupData}$, $\{v_{ij}\}$ (Shift vectors)
\State $\{(C_j, L_j)\}_{j=1}^{T} \gets \text{KMeans}(Z^{\text{source}}, T)$
\State $\text{WarmupData} \gets \{(z_i, l_i)\}$ \Comment{Store labeled data from clustering}
\State $c_j \gets \frac{1}{|C_j|} \sum_{z \in C_j} z \quad \forall j \in \{1, \dots, T\}$ \Comment{Compute centroids}
\State $v_{ij} \gets z_i - c_j \quad \forall z_i \in Z, \forall j \in \{1, \dots, T\}$ \Comment{Compute shift vectors}
\State \Return $\text{WarmupData}$, $\{v_{ij}\}$
\end{algorithmic}
\end{algorithm*}

\begin{algorithm*}[ht]
\caption{Temporal Router Network}
\label{alg:router_network}
\begin{algorithmic}[1]
\Require $\text{WarmupData} (Z, L)= \{(z_i, l_i)\}$, $V = \{v_{ij}\}$, $Z^{*}$, $K$, $\lambda$, $W_g$, $\Theta = \{\Theta_1,...,\Theta_T\}$
\Ensure Prediction $\hat{y}$
\State \textbf{Warmup Router:}
\For{$z_i \in Z$}
    \State $G(z_i) \gets \text{softmax}(W_g z_i)$ \Comment{Compute gating scores}
    \State Optimize $W_g$ using $L_{\text{router}} = -\sum \log l_iG(z_i)$
\EndFor
\State \textbf{Train Router and Experts:}
\For{ $z_i \in Z^{*}$}
    \State $G(z_i) \gets \text{TopK}(\text{softmax}(W_g z_i), K)$ \Comment{Select top-$K$ experts}
    \For{each $j \in \text{TopK}$}
        \State $h_j \gets \Theta_j(z_i) \oplus v_{ij}$ \Comment{Concatenate expert output with shift vector}
        \State $p_j \gets \text{ClassificationLayer}(h_j)$
    \EndFor
    \State $\hat{y} \gets  \frac{1}{|K|} \sum_{j \in \text{TopK}} G_j(z_i) \cdot p_j$ \Comment{Aggregate expert outputs}
\EndFor
\State Compute loss: $\mathcal{L} = CE(y,\hat{y}) + \lambda \mathcal{L}_{\text{aux}}$
\State Update: $W_g, \Theta$
\State \Return Prediction $\hat{y}$
\end{algorithmic}
\end{algorithm*}

\section{Hardware and Software}
The experiments on reciew data are conducted on a device equipped with 2 NVIDIA 4090 GPUs (24GB memory) and an AMD Ryzen 9 7950X CPU, running Ubuntu 22.04. 
The system utilizes PyTorch 2.0 ~\cite{Paszke2019pytorch} alongside HuggingFace Transformers 4.26 ~\cite{wolf2020transformers}. The experiments on legal data are conducted on a machine from iTiger GPU cluster equipped with 8x H100 GPUs, 2x EPYC Genoa 9334 CPUs, and 768GB of RAM. 
The system runs on Linux kernel 5.14.

\section{Data}

We use a public multi-lingual review data ~\cite{hovy2015user} and a legal document data EURLEX~\cite{chalkidis2021multieurlex} for our study.
The review data spans years between 2007 and 2014 and covering four language corpora, Danish, English, French, and German \footnote{The data files are available from the original authors at: https://bitbucket.org/lowlands/release/src/master/WWW2015/data/}. 
The legal data spans the years 1958 to 2016 and covers 23 languages\footnote{The data files are available from the original authors at: \url{https://huggingface.co/datasets/coastalcph/multi_eurlex}}.


\subsection{Data Partition}
To examine temporal effect on the multilingual data, we use the review data corpora (four languages) and partitioned each language corpus into four distinct time domains. For English, French, and German, these domains correspond to the years [2011, 2012, 2013, 2014]. The Danish corpus, due to its longer time span, was divided into [2007-2008, 2009-2010, 2011-2012, 2013-2014].
To ensure comparability across time domains, we downsampled the data in each domain to match the size of the smallest domain. This sampling process was conducted with a fixed random seed (random state=1) for reproducibility.
Within each time domain, we further split the data into training and testing sets, allocating 70\% for training and 30\% for testing. This partition was also performed using a fixed random seed (random state=1) to maintain consistency across experiments.

To evaluate the effectiveness of our proposed method compared to baseline models for time adaptation, we use the data form two domains: the review and the legal data.
For the review data, we re-split the data, holding 20\% of the most recent time period from each corpus as the target time domain, with the remaining data used as the source time domain. For the legal data, we followed the standard train-test split of the EURLEX dataset, using the training split as the source time domain.

\begin{table}[]
\resizebox{1\columnwidth}{!}{
\begin{tabular}{c|cccc}
\textbf{Language}   & \textbf{ISO} & \textbf{Train/Test Docs} & \textbf{Length} & \textbf{I-ratio} \\ \hline \hline
English    & en  & 55,000 / 5,000  & 1200           & 50.11           \\
German     & de  & 55,000 / 5,000  & 1085           & 50.11           \\
French     & fr  & 55,000 / 5,000  & 1280           & 50.11           \\
Italian    & it  & 55,000 / 5,000  & 1210           & 50.11           \\
Spanish    & es  & 52,785 / 5,000  & 1380           & 49.15           \\
Polish     & pl  & 23,197 / 5,000  & 1200           & 37.8            \\
Romanian   & ro  & 15,921 / 5,000  & 1500           & 32.4            \\
Dutch      & nl  & 55,000 / 5,000  & 1230           & 50.11           \\
Greek      & el  & 55,000 / 5,000  & 1230           & 50.11           \\
Hungarian  & hu  & 22,664 / 5,000  & 1120           & 37.25           \\
Portuguese & pt  & 23,188 / 5,000  & 1290           & 49.63           \\
Czech      & cs  & 23,187 / 5,000  & 1170           & 37.73           \\
Swedish    & sv  & 42,490 / 5,000  & 1130           & 46.05           \\
Bulgarian  & bg  & 15,986 / 5,000  & 1480           & 32.22           \\
Danish     & da  & 55,000 / 5,000  & 1080           & 50.11           \\
Finnish    & fi  & 42,497 / 5,000  & 890            & 46.06           \\
Slovak     & sk  & 15,986 / 5,000  & 1180           & 37.9            \\
Lithuanian & lt  & 23,188 / 5,000  & 1070           & 37.96           \\
Croatian   & hr  & 7,944 / 2,500   & 1490           & 26.14           \\
Slovene    & sl  & 23,184 / 5,000  & 1170           & 37.79           \\
Estonian   & et  & 23,126 / 5,000  & 950            & 37.56           \\
Latvian    & lv  & 23,188 / 5,000  & 1080           & 37.75           \\
Maltese    & mt  & 17,521 / 5,000  & 1250           & 32.47          
\end{tabular}}
\caption{EURLEX statics per language: ISO language code; number of documents in train, development, and test splits; average document length (rounded up); and label imbalance ratio. The imbalance ratio is calculated as the frequency of the most common label divided by the least common label across the entire dataset. }
\label{tab: eurlex}
\end{table}

\begin{table*}[h]
    \centering
    \resizebox{1\textwidth}{!}{
    \begin{tabular}{lcccccccccccc}
        \hline
        \textbf{Method} & \multicolumn{3}{c}{\textbf{English}} & \multicolumn{3}{c}{\textbf{French}} & \multicolumn{3}{c}{\textbf{German}} & \multicolumn{3}{c}{\textbf{Danish}} \\
        
         & F1-ma & AUC & Fair ↓ & F1-ma & AUC & Fair ↓ & F1-ma & AUC & Fair ↓ & F1-ma & AUC & Fair ↓ \\
        \hline
        Best-baseline & 47.48 & 78.51 & 2.23 & 41.21 & 75.94 & 1.99 & 38.09 & 74.78 & 2.17 & 47.45 & 81.71 & 2.14 \\
        Source Model  & 48.63 & 78.17 & 2.21 & 34.86 & 72.19 & 1.96 & 33.64 & 74.82 & 2.16 & 49.51 & 82.32 & 2.16 \\
        MoTE (ours)   & \textbf{49.29} & \textbf{79.81} & 2.18 & \textbf{42.06} & \textbf{76.68} & 1.98 & \textbf{40.22} & \textbf{75.66} & 2.16 & \textbf{50.34} & \textbf{83.61} & \textbf{2.05} \\
    \hline
    \end{tabular}
    }
    \caption{Performance on distant period testing across the four languages corpus in the review data.}
    \label{tab:distant_period_results}
\end{table*}

\section{Experimental Details}
\label{sec:appendix:implementation}


We use \textit{XLM-RoBERTa-base}~\cite{xlm-roberta} as the base model and feature encoder in our temporal effect analysis experiments. 
We use the training data of each time domain to fine-tune the base model. 
For review data, each model is trained 10 epoches and configured with a maximum token length of 128, a learning rate of 3e-5, a batch size of 32 per device, and gradient accumulation steps set to 2. For legal data, each model follow same setting except models are trained 5 epoches and maximum token length set to 521 due to longer text length.
We save the checkpoints of the last epoch for testing performance on both the same and the other time domain's test data (the in-time-domain and cross-time-domain evaluation)
One of the time domain is chosen as source domain by the largest performance disparity in cross-time-domain evaluations, and the corresponding model is called the \textit{\textbf{Source Model}}. We compare our proposed method with the source model and baseline methods.

All experiments are run with three random seeds: 41, 42 and 43, and we report the average performance in our paper.

\subsection{Baseline Methods}
\subsubsection{Self Labeling}
Following the paper's~\cite{agarwal2022temporal} setting.
We use the saved source model, as mentioned in last section, to label the test data in the target domain. 
The silver-labeled target data, along with the gold-labeled source data, is then used to fine-tune \textit{XLM-RoBERTa-base}~\cite{xlm-roberta} model with the same training configurations as the source model: 10 epochs, a learning rate of 3e-5, a batch size of 32 per device, and gradient accumulation steps set to 2, and a maximum token length of 128 for the review data and 512 for the legal documents.
\subsubsection{Anti-CF}
We implement the Anti-CF~\cite{su2023beware} framework in classification task. We maintain the same optimization goal and hyper parameters as described in their paper (learning rate = 5e-5 and parameter $\alpha$ = 0.2 for the loss). We apply Anti-CF adaptation to the source model and test the adapted model's performance on the target domain test data.
\subsubsection{ChronosLex} 
For ChronosLex~\cite{santosh2024chronoslex}, we follow their setting and train the model sequentially using data from chronologically ordered 4 time periods. 
Each iteration initializes the model with weights from the previous time step and fine-tunes it on data from the current period before moving to the next, while maintaining the model architecture and loss function.

\subsection{Main Experiments}

In our Mixture of Temporal Experts (MoTE) adaptation method, we use source model as the base model and feature encoder.
We build a temporal routing network including a router and a mixture of experts on the base model.
We apply a TopK gating to the router, with the K set to 2. 
For the training process, we use the AdamW optimizer with learning rate of 1e-4 for router warm-up and experts training.
We use cross-enropy loss in the warmup process, and add an auxiliary load-balancing loss~\cite{shazeer2017outrageously} with a 0.01 weight  following their paper's best setting.
We warmup the router 20 epoches and another 20 epoches train the whole temporal routing network, with batch size of per device set to 32, and gradient accumulation steps set to 2.

\section{Supplementary Experiments Results}

\subsection{Experiments on Distant Period Testing}
We conducted experiments on review data using only the first time-domain’s data, and test it on the last time-domain’s test set.
The results indicate that our method consistently improves performance across various languages and metrics, as shown in Table~\ref{tab:distant_period_results}.

As shown in Table~\ref{tab:distant_period_results}, our method consistently outperforms the best-baseline (ChronosLex) and the source model across all evaluation metrics. For example, in English, MoTE achieves an F1-ma of 49.29, surpassing the best-baseline (47.48) and the source model (48.63). The AUC score also improves to 79.81 compared to 78.51 from the baseline. Notably, MoTE maintains a competitive fairness score of 2.18. 

In French, our method shows significant improvements, with an F1-ma of 42.06 compared to 41.21 from the baseline and 34.86 from the source model. Additionally, the AUC improves to 76.68 while keeping the fairness score at 1.98. Similar performance gains are observed across German and Danish, demonstrating the robustness of our method in distant period settings.

\end{document}